%% file: main.tex
\documentclass{article}

\usepackage[nonatbib,preprint]{neurips_2025}

\usepackage[utf8]{inputenc} 
\usepackage[T1]{fontenc}    
\usepackage{hyperref}       
\usepackage{url}            
\usepackage{booktabs}       
\usepackage{amsfonts}       
\usepackage{nicefrac}       
\usepackage{microtype}      
\usepackage{xcolor}         
\usepackage{graphicx}
\usepackage{amsmath}
\usepackage{algorithm}
\usepackage{algpseudocode}
\MakeRobust{\Call}
\usepackage{multirow}

\title{CycleSAM: Few-Shot Surgical Scene Segmentation with Cycle- and Scene-Consistent Feature Matching}

\author{%
Aditya Murali$^{1,2}$ \quad Faradihba Zarin$^{1,2}$ \quad Adrien Meyer$^{1,2}$ Pietro Mascagni$^{2,3}$ \\ \quad \textbf{Didier Mutter}$^{1,2}$ \textbf{Nicolas Padoy}$^{1,2}$\\\\
$^1$University of Strasbourg, CNRS, INSERM, ICube, UMR7357, Strasbourg, France \\ $^2$IHU Strasbourg, Strasbourg, France \\ $^3$Fondazione Policlinico Universitario A. Gemelli IRCCS, Rome, Italy\\
\texttt{\{murali,fzarin,ameyer1,d.mutter,npadoy\}@unistra.fr}\\
\texttt{pietro.mascagni@ihu-strasbourg.eu}
}

\begin{document}

\maketitle

\input{sec/0_abstract}    
\input{sec/1_intro}
\input{sec/2_related_work}
\input{sec/3_methods}
\input{sec/4_results}
\input{sec/5_conclusion}

\bibliographystyle{splncs04}
\bibliography{main}

\end{document}

%% file: sec/0_abstract.tex
\begin{abstract}
Surgical image segmentation is highly challenging, primarily due to scarcity of annotated data. Generalist prompted segmentation models like the Segment-Anything Model (SAM) can help tackle this task, but because they require image-specific visual prompts for effective performance, their use is limited to improving data annotation efficiency.
Recent approaches extend SAM to automatic segmentation  by using a few labeled reference images to predict point prompts; however, they rely on feature matching pipelines that lack robustness to out-of-domain data like surgical images.
To tackle this problem, we introduce CycleSAM, an improved visual prompt learning approach that employs a data-efficient training phase and enforces a series of soft constraints to produce high-quality feature similarity maps. 
CycleSAM label-efficiently addresses domain gap by leveraging surgery-specific self-supervised feature extractors, then adapts the resulting features through a short parameter-efficient training stage, enabling it to produce informative similarity maps.
CycleSAM further filters the similarity maps with a series of consistency constraints before robustly sampling diverse point prompts for each object instance. 
In our experiments on four diverse surgical datasets, we find that CycleSAM outperforms existing few-shot SAM approaches by a factor of 2-4x in both 1-shot and 5-shot settings, while also achieving strong performance gains over traditional linear probing, parameter-efficient adaptation, and pseudo-labeling methods.
\end{abstract}

%% file: sec/1_intro.tex
\section{Introduction}
\label{sec:intro}

Surgical video understanding is critical to numerous applications in digital surgery, including workflow analysis~\cite{twinanda2016endonet,zisimopoulos2018deepphase,jin2017sv,ramesh2021multi,FunkeBOBWS19,nwoye2022rendezvous,sharma2023surgical,lin2022instrument}, surgical skill assessment~\cite{sharma2025early,wu2021cross,liu2021towards,funke2019video}, and intra-operative assistance tools~\cite{murali2023latent,li2023automated,pavone2025introducing,khalid2023use}.
Surgical image segmentation is a particularly valuable task for downstream applications such as action triplet recognition and critical view of safety assessment~\cite{murali2023latent,sharma2023surgical,khalid2023use}; however, training specialist segmentation models requires high-quality annotated datasets, which are quite scarce in surgical computer vision as they require complex annotation protocols, trained expert clinicians, and often mediation to resolve disagreements~\cite{mascagni2021surgical,carstens2023dresden}.

\begin{figure}[t]
    \centering
    \includegraphics[width=\textwidth]{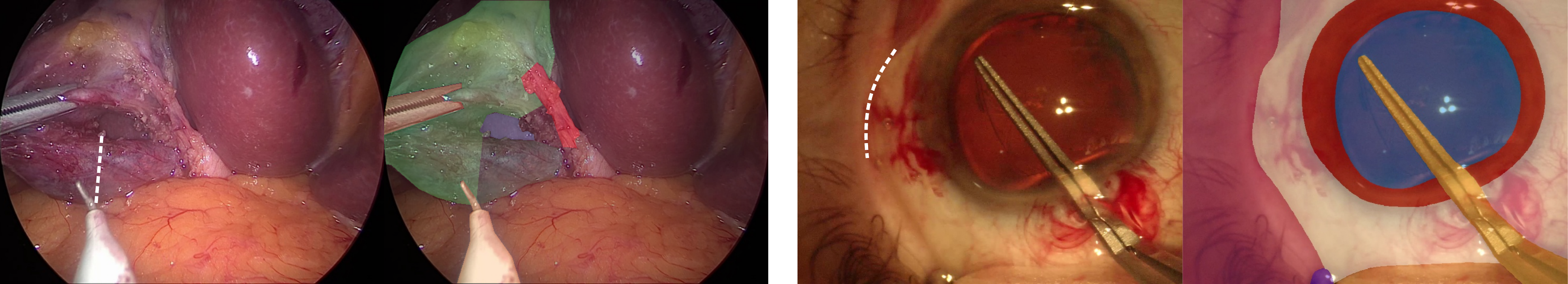}
    \caption{Visually ambiguous object boundaries (dotted white lines) in surgical images. On the left (Endoscapes-Seg50), we see that the gallbladder (light green) and cystic duct (dark green) are similar in appearance and the boundary is defined as where the hepatocystic triangle (blue) begins. On the right (CaDIS), we see a similar boundary between the skin of the eyelid (pink) and the cornea (white).}
    \label{fig:visual_ambiguity}
\end{figure}


Generalist segmentation models such as the Segment Anything Model (SAM) can help circumvent this annotation bottleneck as they provide strong off-the-shelf performance even on surgical data.
However, because they require precise image-specific visual prompts like points or bounding boxes, they are limited to interactive segmentation applications.
Recent works have proposed extensions to SAM for few-shot segmentation based on predicting these visual prompts: Personalize-SAM (PerSAM)~\cite{zhang2023personalize} used labeled reference image to compute a target feature then predicted point prompts in query images using feature matching.
Later approaches like Matcher~\cite{liumatcher} and GF-SAM~\cite{zhang2024bridge} improved PerSAM by using self-supervised feature extraction, improved point prompt selection, and sophisticated mask filtering.
Yet, surgical images pose several unique challenges for these approaches: (1) they present substantial domain shift, (2) they contain anatomical boundaries that are defined heuristically rather than by clear visual indications (see Fig. \ref{fig:visual_ambiguity}), and (3) they require more descriptive visual prompts (e.g. bounding boxes, numerous points) for effective segmentation.
Because these existing methods rely on high-quality feature matching, model each object class independently, and decode masks from single point prompts, they struggle to overcome these challenges (see Sec. \ref{sec:results}).



To tackle these limitations, we introduce \textbf{CycleSAM}, an improved few-shot SAM adaptation framework specifically tailored to surgical image segmentation.
CycleSAM first addresses the domain gap problem by integrating self-supervised feature extractors that are pretrained on unlabeled surgical images, thereby remaining label-efficient.
Then, to produce higher quality feature similarity maps, CycleSAM trains a linear adapter layer and learns per-class target features, enabling it to more robustly capture each object's characteristics and avoid reference feature computation at inference-time. 
To model inter-object relationships, CycleSAM filters the produced feature similarity maps based on spatial cycle-consistency and semantic-consistency heuristics.
Finally, using these similarity maps, CycleSAM separates object instances and selects multiple foreground and background point prompts.



We evaluate CycleSAM on four public surgical scene segmentation datasets that cover three different surgical procedures and include various anatomical structures and tools.
We find that existing SAM-based few-shot segmentation approaches perform poorly in these settings, even failing to reach the performance of a specialist Mask2Former model trained on just a few images.
Meanwhile, thanks to its multitude of improvements, CycleSAM vastly outperforms baseline approaches, both SAM-based and specialist, in both 1-shot and 5-shot settings.
Last but not least, we show that self-training a Mask2Former model using CycleSAM-generated pseudo-masks yields up to 85\% of fully supervised performance—highlighting CycleSAM's potential to drastically reduce annotation requirements.

In summary, our contributions are as follows:
\begin{enumerate}
  \item We propose \textbf{CycleSAM}, a SAM-adaptation approach for surgical image segmentation that leverages domain-specific SSL features, a lightweight trainable feature adapter, and consistency masking to produce high-fidelity feature similarity maps.
  \item We introduce a robust point selection module that samples multiple points per object instance, enabling improved mask decoding for complex anatomical structures.
  \item We vastly outperform state-of-the-art SAM-based methods and specialist models for few-shot segmentation, reaching up to 85\% of fully-supervised performance in the 5-shot setting.
\end{enumerate}

\begin{figure*}[t]
    \centering
    \includegraphics[width=\textwidth]{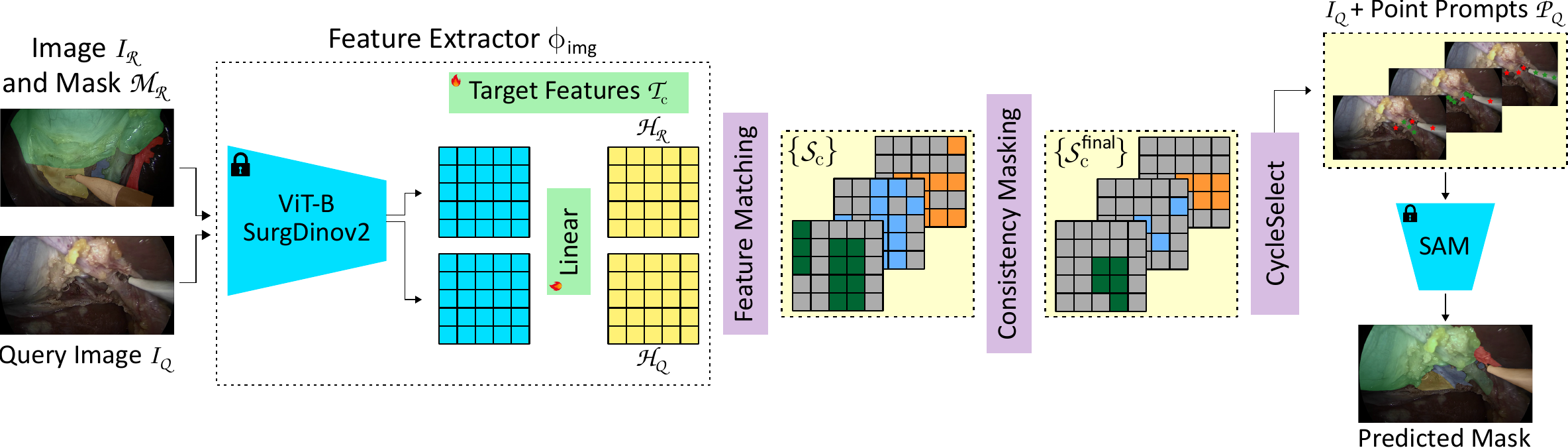}
    \caption{CycleSAM Overview. We first extract feature maps from a reference ($\mathcal{I}_{\mathcal{R}}$) and query image ($\mathcal{I}_{\mathcal{Q}}$). Then, a set of target features are used to compute per-class similarity maps in the query image. These similarity maps are then filtered and passed to CycleSelect, which separates instances of the same class and selects foreground and background point prompts $\mathcal{P}_{\mathcal{Q}}$ for each instance. These prompts are passed to SAM along with the query image to produce segmentation masks.}
    \label{fig:cyclesam_arch}
\end{figure*}

%% file: sec/2_related_work.tex
\section{Related Work}
\label{sec:related_work}

\subsection{Few-Shot Segmentation}
Most traditional few-shot segmentation (FSS) works rely on meta-learning~\cite{catalano2023few,dong2018few,shaban2017one,zhang2021few,Wang_2019_ICCV,Lang_2022_CVPR,Li_2021_CVPR}, requiring training on a vast but diverse set of labeled images.
Initial methods aimed to predict image-specific parameters to condition a segmentation model at test-time~\cite{shaban2017one,rakelly2018few,lu2021simpler,zhang2021rich}.
Other works focused on generating image-mask pairs in order to scale training datasets from a few samples~\cite{saha2022ganorcon,tritrong2021repurposing,zhang2021datasetgan}.
Later methods shifted towards prototype-learning~\cite{dong2018few,Wang_2019_ICCV,zhang2021few,Lang_2022_CVPR,Li_2021_CVPR,Min_2021_ICCV}, learning object target features from reference images and using feature matching to produce segmentation masks.
Most recently, SegGPT~\cite{wang2023seggpt} and Painter~\cite{Wang_2023_CVPR} achieve strong performance with visual in-context learning, modeling segmentation as image generation conditioned on prompt images. 
Yet, these methods require large-scale pre-training on diverse labeled data, which is prohibitive in the surgical domain.

\subsection{Few-Shot Adaptation of SAM}

SAM~\cite{kirillov2023segment} has revolutionized interactive segmentation through extremely large-scale pretraining on high-quality data.
Many recent works have thus focused on adapting the Segment Anything Model (SAM) for few-shot segmentation.
One category of approach is automatic prompt generation~\cite{zhang2023personalize,feng2024boosting,Xie_2024_WACV,wu2023self,zhang2024bridge,xu2024sammpa}.
Initial approaches simply used SAM for refined mask generation, starting with predictions from FSS methods~\cite{feng2024boosting,wu2023self}.
Others use SAM as the segmentation model with a meta-learning FSS framework~\cite{sun2024vrp,xu2024sammpa,leng2024self}.
Personalize-SAM~\cite{zhang2023personalize}, meanwhile, took inspiration from prototype learning FSS approaches to directly identify point prompts in a query image with dense feature matching from a reference object feature.
Matcher~\cite{liumatcher} builds on Personalize-SAM, using DinoV2 feature maps to compute feature similarity, filters point prompts with a bidirectional matching constraint, and filters predicted masks through heuristics to assess mask quality and coverage.
GF-SAM~\cite{zhang2024bridge} reduces Matcher's reliance on heuristics, introducing a graph-based algorithm to filter point prompts and separate object instances, simultaneously improving inference efficiency.

Other works train SAM adapters~\cite{paranjape2023adaptivesam,yue_surgicalsam,xiao2024cat,wu2023medical,wang2023mathrm,cheng2023sam,hu2023efficiently,wu2025augmenting,liusurgical,matasyoh2024samsurg} for surgical data.
While most of these works rely on manual prompting at test-time, a few tackle automatic segmentation~\cite{paranjape2023adaptivesam,yue_surgicalsam,sivakumar2025sasvi}:
AdaptiveSAM~\cite{paranjape2023adaptivesam}  learns prompts from text input, while SurgicalSAM~\cite{yue_surgicalsam} trains a prompt encoder from scratch with a learnable embedding for each target class, additionally finetuning SAM's mask decoder.
While effective, these latter approaches are untested in few-shot settings.

\subsection{Self-Supervised Learning for Surgery}
Because unlabeled surgical video data is abundant, many works have turned to self-supervised learning to build surgical-domain specific feature extractors~\cite{alapatt2024jumpstarting,ramesh2023dissecting,wang2023foundation,batic2024endovit,tian2025endomamba,dermyer2025endodino}.
The self-supervised features from such approaches can be leveraged for zero-shot and few-shot segmentation, either through traditional finetuning or specialized approaches~\cite{wang2023cut,zou2022spot,liu2024rethinking,kalapos2022self,ouyang2022self}.
CutLER~\cite{wang2023cut}, for example, enables few-shot instance segmentation by self-training a segmentation model on pseudo-masks generated from self-supervised features.
Though CutLER is versatile, with applications ranging from label-efficient training of SAM-style models~\cite{wang2025segment} to surgical image segmentation~\cite{koksal2024surgivid}, it produces class-agnostic pseudo-masks, hindering few-shot specialist model training.


%% file: sec/3_methods.tex
\section{Methods}
\label{sec:methods}

In this section, we describe \textbf{CycleSAM}, which takes a set of reference (training) images $\{\mathcal{I}_r, k \in [1...\mathcal{R}]\}$, corresponding reference masks $\{\mathcal{M}_r, k \in [1...\mathcal{R}]\}$, query image $\mathcal{I}_\mathcal{Q}$, an image encoder $\phi_{\text{img}}$, and SAM model $\phi_{\text{SAM}}$ and outputs a binary mask $\mathcal{M}_c$ for each object class $c \in [1...C]$.
CycleSAM contains two modules: CycleSim, which computes per-class patch-wise similarity maps $\mathcal{S}_c$ using a set of learnable per-class target features $\mathcal{T}_c$ for the query image $\mathcal{I}_\mathcal{Q}$, and CycleSelect, which outputs $N$ sets of point prompts $\mathcal{P}^{n}_{\mathcal{Q}}$ for the query image using $\mathcal{S}$, where $N$ is the number of predicted instances.

\subsection{CycleSim}
\label{subsec:cyclesim}

CycleSim is based on the key assumption that patch features localized to an object class $c$ should be similar across images.
Based on this assumption, for each class $c$, it computes a similarity map $\mathcal{S}_c$ by comparing each patch feature of a query image to a class-specific target feature $t_c$. 
This similarity map $\mathcal{S}_c$ is then filtered via a series of soft constraints that encourage consistency among the similarity maps for different objects in an image (e.g. each image patch should match to a single object class).
The resulting final similarity map $\mathcal{S}^{\text{final}}_c$ is passed to CycleSelect to select prompts for SAM.

\textbf{Learnable Target Features.} 
Previous few-shot SAM adaptations~\cite{zhang2023personalize,liumatcher,zhang2024bridge} use each reference image-mask pair to compute a target feature for each class; however, this strategy does not scale well when we have access to multiple reference images (e.g. 5), as we either need to maintain multiple target features or use heuristics (mean, max) to obtain a single target feature.
CycleSim instead \textit{learns} target features $\mathcal{T}_c \in \mathbb{R}^{D}$ for each class.
Then, at test-time, the learned target features are used directly to compute the per-class similarity maps $\mathcal{S}_c$; this inference pipeline can therefore scale to arbitrary numbers of reference images.

\textbf{Feature Extraction and Adaptation.} 
High-quality visual features are critical for robust feature matching, but many off-the-shelf feature extractors, including SAM's ViT backbone, produce noisy feature maps on surgical images due to substantial domain shift.
To address this challenge, given the abundance of unlabeled surgical images relative to annotated ones, we train dataset-specific DinoV2 feature extractors through self-supervised fine-tuning, as pioneered by recent works~\cite{ramesh2023dissecting,alapatt2024jumpstarting}\footnote{This is especially true in surgical datasets that originate from videos where most frames are unlabeled for dense tasks like segmentation.}.

While custom DinoV2 feature extractors effectively address domain shift, SSL-pretrained visual feature extractors naturally struggle to delineate heuristically-defined or dataset-specific anatomical structures/regions (see Fig. 1).
Training-free approaches like Matcher and GF-SAM rely on SAM's multi-granularity predictions and sophisticated mask post-processing pipelines to tackle these scenarios; in the surgical domain however, we posit that a small training stage can be a more generalizable, efficient, and effective alternative.
To this end, we adapt the DinoV2 image features with a lightweight linear projector.
Our final feature extractor is therefore a composition of a frozen DinoV2 backbone ($\phi_{\text{dinov2}}$) and a trainable linear projector $\phi_{\text{projector}}$:

\begin{equation}
    \phi_{\text{img}}(x) = \phi_{\text{projector}}(\phi_{\text{dinov2}}(x)).
\end{equation}

Using $\phi_{\text{img}}$ and the target features $\mathcal{T}_c$, we can compute the initial per-class similarity maps $\mathcal{S}_c$ by computing an element-wise cosine similarity between the query image features and target features:

\begin{equation}
\begin{gathered}
    \mathcal{H}_\mathcal{Q} = \phi_{\text{img}}\left(\mathcal{I}_\mathcal{Q}\right), \\
    \mathcal{S}_c = \frac{\mathcal{H}_\mathcal{Q} \odot \mathcal{T}_c}{||\mathcal{H}_\mathcal{Q}|| \cdot ||\mathcal{T}_c||},
\end{gathered}
\end{equation}
where $\mathcal{H}_\mathcal{Q} \in \mathbb{R}^{D \times H^{\text{feat}} \times W^{\text{feat}}}$ and $\mathcal{S}_c \in \mathbb{R}^{H^{\text{feat}} \times W^{\text{feat}}}$.

\noindent \textbf{Training.} We train $\phi_{\text{projector}}$ and the learnable target features $\mathcal{T}_c$ using the labeled reference images $\mathcal{I}_\mathcal{R}$.
Specifically, for each object class $c$, we extract foreground patch features $\mathcal{H}_{\text{fg}}^{c}$, then pass them to a patch-level contrastive loss, $\mathcal{L}_{\text{feat\_sim}}$, along with the target features $\mathcal{T}_c$.
$\mathcal{L}_{\text{feat\_sim}}$ maximizes the cosine similarity between $\mathcal{T}_c$ and $\mathcal{H}_{\text{fg}}^{c}$,  while minimizing that between $\mathcal{T}_c$ and the foreground features of every other class $\{\mathcal{H}_{\text{fg}}^{\hat{c}},\  \hat{c} \neq c\}$.\footnote{For datasets containing background pixels that do not belong to any object class, we treat the background as an additional instance with its own label and learn a separate target feature for it.}
The $C$ loss values are simply averaged at the end, enabling $\mathcal{L}_{\text{feat\_sim}}$ to consider each object class independently of object size.

\subsubsection{Consistency Masking}
While we could directly use $\mathcal{S}_c$ to sample point prompts, this often leads to spurious matches because the feature maps produced by $\phi_{\text{img}}$ are not sufficiently discriminative, particularly at the point granularity.
We therefore filter $\mathcal{S}_c$ with three different masking strategies.

\textbf{Spatial Cycle-Consistency.} 
Several prior works relying on feature similarity~\cite{zhang2021few, hundt2021good,jabri2020walk,lebailly2023cribo,Son_2022_CVPR,liumatcher}, have employed a spatial cycle consistency constraint to filter similarity maps.
Intuitively, such a constraint can help filter out poor matches that result from noisy feature maps and local ambiguities, and is particularly important for dealing with query images that do not contain certain target object classes.

To incorporate this into CycleSim, we modify the bi-directional matching approach of Matcher~\cite{liumatcher}, which first finds the patch features in the query image $\mathcal{I}_\mathcal{Q}$ that most closely match a reference object of class $c$, then enforces these patches to rematch to reference image patches that lie within the mask for class $c$.
Importantly, while Matcher uses this constraint to filter individual point prompts, we aim to produce improved similarity maps $\mathcal{S}$.
Consequently, we start by computing a spatial cycle-consistency mask $\mathcal{M}_{\text{SCC}}^{c}$ in four steps: for each reference image, we (1) extract features, (2) densely compute the cosine similarity between all query and reference features, (3) find the index of the closest reference feature to each query feature, and (4) set $\mathcal{M}_{\text{SCC}}^{c, r}$ to $0$ if the index of the rematched feature is not part of the reference object mask, otherwise $1$.

We then rescale $\mathcal{M}_{\text{SCC}}^{c}$ to $[-1, 1]$ and combine it with the original similarity matrix $\mathcal{S}^{c}$ into a final cycle-consistent similarity matrix $\mathcal{S}_{\text{SCC}}^{c}$ as follows:
\begin{equation}
    \mathcal{S}_{c}^{\text{SCC}} = \lambda_{\text{SCC}} \mathcal{M}_{c}^{\text{SCC}} + (1 - \lambda_{\text{SCC}}) \mathcal{S}_c,
\end{equation}

where $\lambda_{\text{SCC}}$ is a weighting factor.

\textbf{Semantic Consistency.} 
To handle scenarios where multiple object classes are present in a query image, we independently compute similarity maps $\mathcal{S}_c$ for each class. However, this approach lacks semantic awareness, as it treats each class in isolation.
To incorporate class-awareness into the similarity maps, we also integrate a semantic consistency constraint, computing a semantic consistency mask $\mathcal{M}_{c}^{\text{sem}}$ as the softmax across the classes at each spatial location in the similarity map.
We then rescale the output to the range $[-1, 1]$ and compute a weighted sum with the cycle-consistency-masked similarity map to obtain our semantically consistent similarity map $\mathcal{S}^{\text{SC}}$.
Concretely:
\begin{equation}
\begin{gathered}
    \mathcal{M}^{\text{SC}} = softmax_{c}\left(\mathcal{S}^{\text{SCC}}\right) \\
    \mathcal{S}^{\text{SC}} = \lambda_{\text{SC}} \mathcal{M}_c^{\text{SC}} + (1 - \lambda_{\text{SC}}) \mathcal{S}_c^{\text{SCC}}
\end{gathered}
\end{equation}


\textbf{Auxiliary Semantic Segmentation.} 
In addition to these masking strategies, we note that the overall similarity map $\mathcal{S}$ has effectively the same structure as a semantic segmentation mask.
To further improve $\mathcal{S}$, we introduce an auxiliary semantic segmentation head, $\phi_{\text{semseg}}$, which directly predicts a semantic segmentation mask from the query image features:
\begin{equation}
    \mathcal{M}_{\text{semseg}} = \phi_{\text{semseg}}(\mathcal{H}_\mathcal{Q}),
\end{equation}
where $\mathcal{M}_{\text{semseg}} \in \mathbb{R}^{C \times H^{\text{feat}} \times W^{\text{feat}}}$.

As before, we integrate the two using a weighted sum, yielding the final similarity map $\mathcal{S}^{\text{final}}$:

\begin{equation}
    \mathcal{S}^{\text{final}} = \lambda_{\text{semseg}} \mathcal{M}_{\text{semseg}} + (1 - \lambda_{\text{semseg}}) \mathcal{S}^{\text{SC}}
\end{equation}

where $\lambda_{\text{semseg}}$ controls the contribution of the auxiliary segmentation prediction.

\subsection{CycleSelect}

Prior works~\cite{zhang2023personalize,liumatcher,zhang2024bridge} use a single foreground point prompt to prompt SAM; however, more sophisticated prompting strategies can significantly improve mask decoding performance, particularly in surgical images.
CycleSelect improves point prompt selection with four key strategies.

\noindent \textbf{Similarity Connected Component Separation.}
CycleSim outputs per-class similarity maps, but we require an explicit strategy to separate multiple instances of the same class, where present.
Consequently, we begin by separating the connected components within each similarity map.
For each object class $c$, we first threshold the similarity map $\mathcal{S}_c$ using a predefined threshold $\lambda_{\text{sim}}$ to produce a binary mask.
Then, we select the $N$ largest connected components, each representing an instance-specific similarity map $\mathcal{S}_i$.

\noindent \textbf{Diverse Foreground Point Selection.} To improve foreground point selection, we iteratively sample $P$ diverse points from each instance similarity map $\mathcal{S}_i$. Selecting only the highest-similarity points would result in redundant prompts, so we instead enforce spatial diversity using a distance transform approach:
(1) compute the spatial distance from each foreground pixel to the nearest background pixel.
(2) select the pixel with the highest distance and add it to the foreground prompt set.
(3) update the background set to include the newly selected point and recompute distances.
(4) repeat until $P$ diverse foreground points are selected.

\noindent \textbf{Boundary Background Point Selection.} To select informative negative points, we explicitly sample $B$ points near the object boundary. We define a boundary region for each instance as:
\begin{equation}
    \mathcal{S}_i^{\text{boundary}} = \mathcal{D}(\mathcal{S}_i) - \mathcal{S}_i,
\end{equation}
where $\mathcal{D}(\mathcal{S}_i)$ represents a morphological dilation of $\mathcal{S}_i$. We then apply the same distance transform-based sampling strategy to select $N_{\text{boundary}}$ background points from this boundary region.

Altogether, we end up with $P + B$ total point prompts for each instance $i$; we pass the prompts for each instance $\mathcal{P}_{\mathcal{Q}}^i$ to $\phi_{\text{SAM}}$ to produce the final masks $\mathcal{M}_{\mathcal{Q}}^i$.


%% file: sec/4_results.tex
\section{Results}
\label{sec:results}


\subsection{Experimental Setup}

We evaluate our approach on four publicly available surgical datasets: \textbf{EndoscapesSeg50}~\cite{murali2023endoscapes}, \textbf{CaDIS}~\cite{grammatikopoulou2021cadis}, \textbf{CholecSeg8k}~\cite{hong2020cholecseg8k}, and \textbf{HyperKvasir}~\cite{borgli2020hyperkvasir}. These datasets span three different surgical procedures and range from multi-class multi-instance segmentation (EndoscapesSeg50, CaDIS) and full scene segmentation (CaDIS) to binary segmentation (HyperKvasir).

To evaluate few-shot generalization, we conduct experiments in both \textbf{1-shot} and \textbf{5-shot} settings. For each setting, we randomly sample 3 folds of annotated images from the training set of each dataset. When sampling, we prioritize images with all of the ground truth classes present, and in the absence of such images, select a set of images that together span all object classes.\footnote{For example, in the 1-shot setting on CholecSeg8k, we sample two images to ensure full class coverage.}
We also sample only one image per video to ensure intra-set diversity.

%

\subsection{Main Results}

\begin{table*}[t]
\caption{Performance (Mean Segmentation mAP $\pm$ Std Dev) of Various Methods for 1-shot and 5-shot Instance Segmentation. Results are divided into 4 categories: (1) Self-Supervised Learning Based Approaches, (2) SAM Adaptation Baseline Approaches, (3) Our Proposed Method, (4) SAM Manual Prompting Performance, and (5) Fully-Supervised Performance. The latter two groups are included to represent ceiling performances and aid in the analysis of few-shot model performance.}
\label{table:main_exps}
\centering
\renewcommand{\arraystretch}{1.2}  
\resizebox{\textwidth}{!}{%
\begin{tabular}{ccccccccc}
\multirow{3}{*}{\textbf{Method}}
 & \multicolumn{2}{c}{\textbf{Endoscapes-Seg50}} & \multicolumn{2}{c}{\textbf{CaDIS}} & \multicolumn{2}{c}{\textbf{CholecSeg8k}} & \multicolumn{2}{c}{\textbf{HyperKvasir}} \\
 & \textbf{1 Shot} & \textbf{5 Shot} & \textbf{1 Shot} & \textbf{5 Shot} & \textbf{1 Shot} & \textbf{5 Shot} & \textbf{1 Shot} & \textbf{5 Shot} \\ \hline
CutLER & 2.1 $\pm$ 0.3 & 6.7 $\pm$ 1.0 & 12.3 $\pm$ 3.1 & 22.6 $\pm$ 1.4 & 9.5 $\pm$ 0.2 & 11.8 $\pm$ 1.3 & 20.1 $\pm$ 9.7 & 35.8 $\pm$ 7.1 \\
DinoV2-M2F & 3.8 $\pm$ 1.9 & 10.5 $\pm$ 0.7 & 14.4 $\pm$ 2.7 & 30.9 $\pm$ 2.5 & 12.6 $\pm$ 0.6 & 16.7 $\pm$ 1.5 & 24.9 $\pm$ 10.8 & 40.4 $\pm$ 5.3 \\ \hline
SelfPromptSAM & 0.8 $\pm$ 0.1 & 1.7 $\pm$ 0.3 & 1.2 $\pm$ 0.3 & 3.9 $\pm$ 0.9 & 1.3 $\pm$ 0.0 & 1.9 $\pm$ 0.1 & 3.6 $\pm$ 0.8 & 5.5 $\pm$ 0.6 \\
SurgicalSAM & 2.4 $\pm$ 0.4 & 5.7 $\pm$ 1.1 & 4.0 $\pm$ 0.5 & 15.7 $\pm$ 2.3 & 5.3 $\pm$ 0.2 & 8.9 $\pm$ 1.2 & 20.3 $\pm$ 1.9 & 31.7 $\pm$ 2.4 \\
PerSAM & 0.8 $\pm$ 0.7 & 1.3 $\pm$ 0.8 & 12.2 $\pm$ 1.9 & 11.5 $\pm$ 2.6 & 3.7 $\pm$ 1.8 & 3.0 $\pm$ 0.9 & 1.6 $\pm$ 1.9 & 4.6 $\pm$ 3.1 \\
PerSAM-F & 0.6 $\pm$ 0.5 & 1.3 $\pm$ 0.9 & 11.4 $\pm$ 1.8 & 11.3 $\pm$ 2.5 & 3.5 $\pm$ 1.6 & 2.9 $\pm$ 0.6 & 1.6 $\pm$ 1.9 & 5.6 $\pm$ 3.5 \\
Matcher & 3.8 $\pm$ 0.9 & 4.1 $\pm$ 0.6 & \multicolumn{1}{l}{16.1 $\pm$ 0.5} & \multicolumn{1}{l}{18.9 $\pm$ 0.8} & \multicolumn{1}{l}{11.0 $\pm$ 0.7} & \multicolumn{1}{l}{11.1 $\pm$ 0.2} & 12.0 $\pm$ 8.9 & \multicolumn{1}{l}{22.3 $\pm$ 0.6} \\
GF-SAM & 2.9 $\pm$ 0.5 & 2.6 $\pm$ 0.4 & 11.5 $\pm$ 2.3 & \multicolumn{1}{l}{12.9 $\pm$ 0.4} & \multicolumn{1}{l}{5.8 $\pm$ 2.5} & \multicolumn{1}{l}{6.3 $\pm$ 1.0} & 7.8 $\pm$ 3.4 & \multicolumn{1}{l}{23.0 $\pm$ 2.0} \\ \hline
CycleSAM & 11.0 $\pm$ 2.1 & 13.2 $\pm$ 0.7 & 22.0 $\pm$ 2.4 & 28.7 $\pm$ 0.7 & 16.8 $\pm$ 1.0 & 16.9 $\pm$ 3.1 & \textbf{35.3 $\pm$ 5.9} & 42.5 $\pm$ 1.2 \\
CycleSAM-DF & \textbf{12.7 $\pm$ 1.1} & \textbf{15.9 $\pm$ 0.8} & \textbf{28.6 $\pm$ 2.8} & \textbf{37.4 $\pm$ 2.0} & \textbf{20.3 $\pm$ 2.8} & \textbf{22.4 $\pm$ 1.6} & 31.1 $\pm$ 7.5 & \textbf{42.6 $\pm$ 5.1} \\ \hline
SAM 1 Point & \multicolumn{2}{c}{13.1} & \multicolumn{2}{c}{29.4} & \multicolumn{2}{c}{16.5} & \multicolumn{2}{c}{22.1} \\
SAM 3 Point & \multicolumn{2}{c}{16.5} & \multicolumn{2}{c}{34.9} & \multicolumn{2}{c}{21.9} & \multicolumn{2}{c}{27.1} \\
\begin{tabular}[c]{@{}c@{}}SAM 3 Pos., \\3 Neg. Points \end{tabular} & \multicolumn{2}{c}{19.9} & \multicolumn{2}{c}{40.5} & \multicolumn{2}{c}{27.2} & \multicolumn{2}{c}{46.3} \\
SAM Box & \multicolumn{2}{c}{35.1} & \multicolumn{2}{c}{50.6} & \multicolumn{2}{c}{44.9} & \multicolumn{2}{c}{75.1} \\ \hline
\begin{tabular}[c]{@{}c@{}}Fully Supervised\\ DinoV2-M2F\end{tabular} & \multicolumn{2}{c}{26.5} & \multicolumn{2}{c}{52.8} & \multicolumn{2}{c}{41.0} & \multicolumn{2}{c}{65.9}
\end{tabular}
}
\end{table*}

Table~\ref{table:main_exps} presents the segmentation performance (mAP) of various methods under 1-shot and 5-shot settings across our four evaluation datasets.
We group the methods into five categories for analysis.

\noindent \textbf{Self-Supervised Learning Approaches.} 
We evaluate two SSL-based methods: (1) a straightforward linear probing baseline, DinoV2-Mask2Former, which consists of a frozen, surgical domain pretrained DinoV2 feature extractor and a Mask2Former head, and (2) CutLER~\cite{wang2023cut}, an unsupervised approach that produces class-agnostic instance segmentation masks using SSL-pretrained feature extractors.
To apply CutLER for few-shot instance segmentation, we generate class-agnostic pseudo-masks for the training set of each dataset, train a DinoV2-Mask2Former model on these pseudo-masks, then finally finetune the Mask2Former head on the labeled subset  (leaving the backbone frozen in both steps).

\noindent \textbf{SAM Adaptation Baselines.}
Existing SAM-based few-shot segmentation methods can be broadly divided into 2 categories: prompt-prediction and adaptation.
PerSAM~\cite{zhang2023personalize} (and its variants), Matcher~\cite{liumatcher}, and GF-SAM~\cite{zhang2024bridge} fall into the former category, using feature similarity to compute prompts for SAM.
SelfPromptSAM~\cite{wu2023self} can be considered a hybrid approach as it first trains a linear layer on top of SAM's ViT backbone to predict a segmentation mask, but then samples prompts from the predicted mask to pass to SAM. 
SurgicalSAM~\cite{yue_surgicalsam} is an adaptation approach that trains a custom prompt encoder using only the target class of each object, also finetuning SAM's mask decoder; it has been shown to achieve strong performance for surgical tool segmentation, but is unexplored for few-shot applications and more complex surgical segmentation tasks.

\noindent \textbf{Our Method.}
We evaluate two variants of our method: CycleSAM, where the underlying SAM model is completely frozen, and CycleSAM-DF, where we additionally finetune the mask decoder of SAM.
For this finetuning stage, following SAM~\cite{kirillov2023segment}, we use a linear combination of dice loss and focal loss, sampling point and box prompts from the ground-truth masks.

\noindent \textbf{Performance Ceilings.}
We report the performance of a manually-prompted SAM and fully-supervised DinoV2-Mask2Former (DinoV2-M2F) to illustrate ceiling performance.
We consider 4 prompting modes for SAM: (1) a single point, (2) 3 foreground points, (3) 3 foreground points and 3 near-boundary background points, and (4) a bounding box.
We sample boundary points with the strategy as CycleSelect, using the ground truth mask in lieu of the similarity map.

\subsection{Comparative Analysis}

CycleSAM and CycleSAM-DF demonstrate substantial improvements across the four evaluation datasets in both 1-shot and 5-shot settings. 
Below, we discuss several key performance trends.

\noindent \textbf{Ceiling Performance.}
The performance trends of manually prompted SAM show that rich prompting is extremely impactful in the surgical domain.
For example, performance increases going from a single foreground point to 3 foreground and 3 near-boundary background points, reiterating that object boundaries are particularly difficult to disambiguate in surgical datasets. 
SAM further improves with bounding box prompts, which are much more informative than points, but equally much more difficult to predict in few-shot settings.

\begin{figure*}[t]
    \centering
    \includegraphics[width=\textwidth]{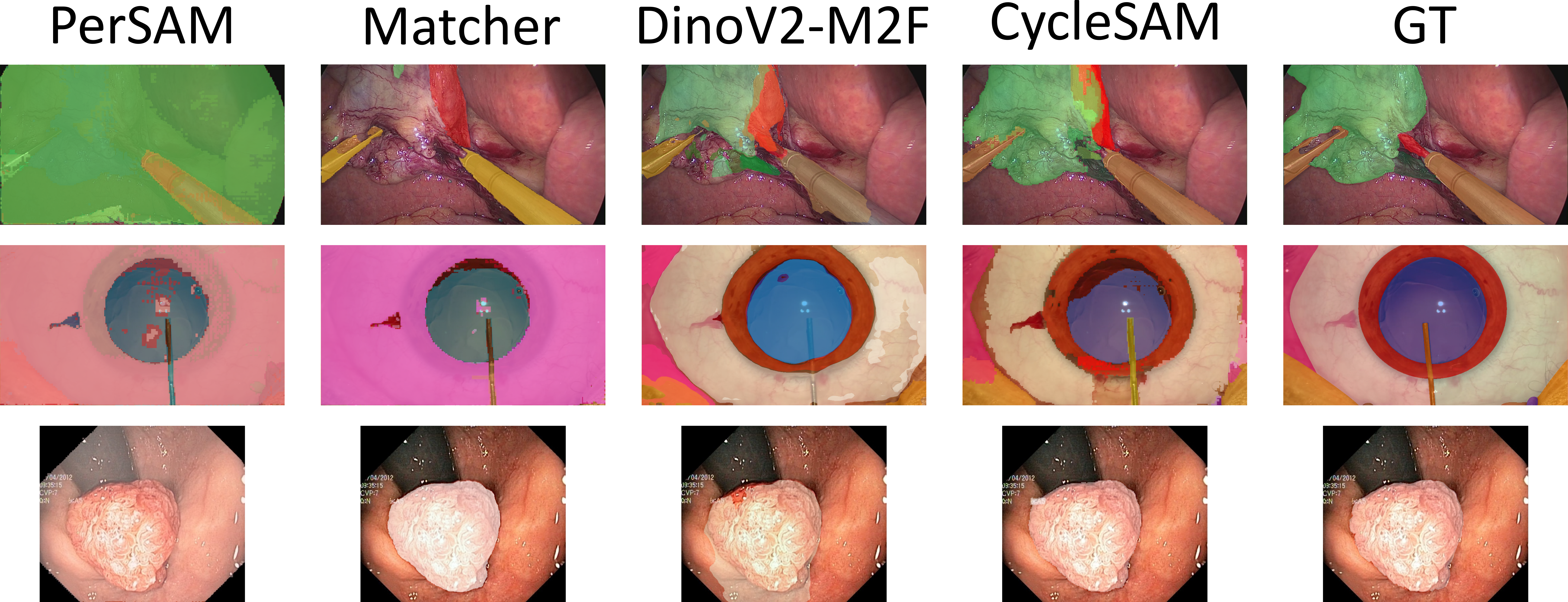}
    \caption{Qualitative results on Endoscapes-Seg50 (top), CaDIS (middle), and HyperKvasir (bottom). PerSAM and Matcher fail to capture most important structures. DinoV2-M2F starts to capture the different objects, but under-segments as it lacks the generalist capabilities of SAM. CycleSAM realizes the benefits of both approaches, yielding consistently good results.}
    \label{fig:qual}
\end{figure*}

\noindent \textbf{Comparison to SAM-based Approaches.}
Existing SAM-based methods struggle on surgical datasets either due to surgery-specific challenges (PerSAM, Matcher, GF-SAM), or because they are not designed for few-shot usage (SelfPromptSAM, SurgicalSAM).
PerSAM is especially poor as it lacks domain-specific feature representations and relies on single point prompts.
We evaluate Matcher and GF-SAM with our surgical domain-pretrained DinoV2 feature extractors, so they achieve modest improvements over PerSAM; yet, they still decode a single point prompt at a time, relying on a mask post-processing pipeline that does not translate well to surgical images.

SelfPromptSAM is quite ineffective as it relies on a supervised training phase to predict rough segmentation masks before sampling bounding box prompts from these masks; in few-shot scenarios, the predicted masks yield inaccurate boxes, limiting performance.
Meanwhile, SurgicalSAM is relatively competitive, especially at 5 shots, as it limits its scope to learning an image-agnostic prompt for each object class, and also includes a mask decoder finetuning stage.
Still, CycleSAM-DF consistently outperforms it by wide margins across datasets, indicating that spatial prompt prediction is a more tractable task in label-constrained settings.

\noindent \textbf{Comparison to SSL-based Approaches.}
Probing the DinoV2 backbone with a Mask2Former head (DinoV2-Mask2Former), is quite effective, outperforming all SAM-based baselines.
CutLER meanwhile performs consistently poorly, likely because it requires complex parameter selection and produces object-agnostic pseudolabels, limiting the effectiveness of the subsequent self-training.

\noindent \textbf{Training vs Training-Free Approaches.}
CycleSAM, CycleSAM-DF, SurgicalSAM, and DinoV2-Mask2Former—all of which involve a training stage—generally outperform training-free approaches like PerSAM, GF-SAM, and Matcher.
These methods also demonstrate a clear benefit from increasing the number of shots: SelfPromptSAM, SurgicalSAM, DinoV2-Mask2Former, and CycleSAM all improve considerably when moving from 1-shot to 5-shot on all 4 evaluation datasets; meanwhile, PerSAM, Matcher, and GF-SAM perform similarly in both 1-shot and 5-shot settings on all datasets except HyperKvasir.
This trend suggests that while surgical images are complex, they are likely also less diverse than natural images, enabling positive gains from training on even a single image.

\begin{table*}[t]
\caption{Ablation Study of CycleSim Components (mean $\pm$ std. segmentation mAP). We start with the base PerSAM and show the impact of adding each of CycleSim's components.}
\label{table:cycle_sim_ablation}
\centering
\resizebox{0.85\textwidth}{!}{%
\begin{tabular}{lcccc}
\multicolumn{1}{c}{\multirow{4}{*}{\textbf{Method}}} \\
\multicolumn{1}{c}{} & \multicolumn{2}{c}{\textbf{ES-50}} & \multicolumn{2}{c}{\textbf{CaDIS}} \\
\multicolumn{1}{c}{} & 1 Shot & 5 Shot & 1 Shot & 5 Shot \\ \hline
PerSAM & 0.8 $\pm$ 0.7 & 1.3 $\pm$ 0.8 & 12.2 $\pm$ 1.9 & 11.5 $\pm$ 2.6 \\
+ CycleSelect & 3.7 $\pm$ 0.8 & 4.3 $\pm$ 0.9 & 9.4 $\pm$ 4.7 & 4.1 $\pm$ 2.4 \\
+ Domain-Specific Pretraining & 5.6 $\pm$ 1.6 & 7.1 $\pm$ 2.1 & 14.5 $\pm$ 5.5 & 15.3 $\pm$ 2.0 \\
+ Feat Sim Adapter & 5.4 $\pm$ 1.2 & 7.4 $\pm$ 1.1 & \multicolumn{1}{l}{16.2 $\pm$ 5.5} & 22.2 $\pm$ 0.6 \\
+ Learnable Target Features & 6.1 $\pm$ 1.9 & 9.6 $\pm$ 1.5 & 14.5 $\pm$ 4.7 & 26.7 $\pm$ 2.2 \\
+ Semantic Consistency & 6.7 $\pm$ 2.1 & 7.2 $\pm$ 1.1 & 18.7 $\pm$ 1.4 & 24.4 $\pm$ 0.6 \\
+ Semantic Segmentation & 9.4 $\pm$ 1.6 & 13.5 $\pm$ 0.4 & 19.3 $\pm$ 0.8 & 26.7 $\pm$ 0.3 \\
\textbf{+ CCD Mask (CycleSAM)} & \textbf{11.0 $\pm$ 2.1} & \textbf{13.2 $\pm$ 0.7} & \textbf{22.0 $\pm$ 2.4} & \textbf{28.7 $\pm$ 0.7} \\
\textbf{+ Decoder Finetuning (CycleSAM-DF)} & \textbf{12.7 $\pm$ 1.1} & \textbf{15.9 $\pm$ 0.8} & \textbf{28.6 $\pm$ 2.8} & \textbf{37.4 $\pm$ 2.0}
\end{tabular}
}
\end{table*}

\begin{table*}[t]
\caption{Ablation Study of CycleSelect (mean $\pm$ std. segmentation mAP). We use the fully functional CycleSim, and start with a single foreground point prompt.}
\label{table:cycle_select_ablation}
\centering
\resizebox{0.85\textwidth}{!}{%
\begin{tabular}{lcccc}
\multicolumn{1}{c}{\multirow{4}{*}{\textbf{Method}}} \\
\multicolumn{1}{c}{} & \multicolumn{2}{c}{\textbf{ES-50}} & \multicolumn{2}{c}{\textbf{CaDIS}} \\
\multicolumn{1}{c}{} & 1 Shot & 5 Shot & 1 Shot & 5 Shot \\ \hline
Single Positive Point & 2.8 $\pm$ 0.5 & 3.1 $\pm$ 0.6 & 16.8 $\pm$ 2.0 & 23.2 $\pm$ 1.3 \\
+ Connected Component Separation & 2.9 $\pm$ 0.5 & 4.1 $\pm$ 0.6 & 17.0 $\pm$ 1.0 & 23.7 $\pm$ 2.0 \\
+ 3 Top Matching FG Points & 3.2 $\pm$ 1.1 & 4.8 $\pm$ 0.1 & 16.7 $\pm$ 1.2 & 23.7 $\pm$ 1.4 \\
+ 3 Evenly Sampled FG Points & 6.3 $\pm$ 2.6 & 9.2 $\pm$ 0.3 & 15.8 $\pm$ 1.7 & 23.2 $\pm$ 0.7 \\
+ 3 Random BG Points & 5.9 $\pm$ 2.1 & 8.9 $\pm$ 0.2 & 12.0 $\pm$ 0.6 & 17.5 $\pm$ 2.3 \\
\textbf{+ 3 Boundary BG Points (CycleSAM)} & \textbf{11.0 $\pm$ 2.1} & \textbf{13.2 $\pm$ 0.7} & \textbf{22.0 $\pm$ 2.4} & \textbf{28.7 $\pm$ 0.7}
\end{tabular}
}
\end{table*}

\begin{table*}[t]
\caption{Self-Training Experiments. We pseudolabel each training set using the CycleSAM-DF model trained on each split of 5-shot labels and report mean and std. of segmentation mAP.}
\label{table:self_training}
\centering
\resizebox{0.85\textwidth}{!}{%
\begin{tabular}{cccc}
\textbf{Dataset} & \textbf{CycleSAM-DF} & \textbf{CycleSAM-ST} & \multicolumn{1}{l}{\textbf{Fully Supervised DinoV2-M2F}} \\ \hline
Endoscapes-Seg50 & 15.9 $\pm$ 0.8 & 18.2 $\pm$ 0.6 & 26.5 \\
CaDIS & 37.4 $\pm$ 2.0 & 45.3 $\pm$ 5.6 & 52.8 \\
CholecSeg8k & 22.4 $\pm$ 1.6 & 31.9 $\pm$ 2.4 & 41.0 \\
HyperKvasir & 42.6 $\pm$ 5.1 & 56.3 $\pm$ 3.8 & 65.9
\end{tabular}}
\end{table*}

\subsection{Ablation and Scaling Studies}

\noindent \textbf{CycleSim Ablation Study.}
Table \ref{table:cycle_sim_ablation} ablates the various components of CycleSim. Using domain-specific pretraining substantially improves CycleSAM's performance, which also explains the performance difference between PerSAM and Matcher/GF-SAM.
Adding the learnable components (feature similarity adapter, target features) yields another sizable performance boost, particularly in the 5-shot settings, which again matches the trends we observed in the main results.
The consistency masks (semantic consistency, semantic segmentation, and spatial cycle consistency), each improve performance across shots and datasets, and together bring an overall relative improvement of $\sim$35\%.
Lastly, finetuning SAM's decoder can further boost performance, particularly in the 5-shot setting.
Importantly, we find that the various components of CycleSim are synergistic, with performance increasing each time a component is added.

\noindent \textbf{CycleSelect Ablation Study.}
Table \ref{table:cycle_select_ablation} ablates the various components of CycleSim.
Prompt selection greatly influences SAM's performance on surgical datasets (see Table \ref{table:main_exps}).
We observe similar trends with CycleSelect: using a single point prompt ($P=1$), even with the fully-functional CycleSim, yields poor results.
Adding connected component separation enables handling multiple object instances of the same class, but does not greatly improve performance.
Using multiple point prompts ($P=3$, $B=3$) can improve performance provided they are effectively sampled; simply selecting the most similar points per instance does not improve performance, whereas evenly sampling points from the thresholded similarity mask yields improvements.
Similarly, sampling three random background points actually harms performance, while using our near-boundary point sampling approach is extremely effective, yielding an average relative improvement of $\sim$37\%.

\noindent \textbf{Self-Training Experiments.} 
We use the 5-shot CycleSAM-DF to generate pseudomasks on each benchmark dataset's training set, then train DinoV2-Mask2Former models on the pseudomasks.
The resulting CycleSAM-ST (Table \ref{table:self_training}) achieves between 68\% and 80\% of fully-supervised performance.

%% file: sec/5_conclusion.tex
\section{Conclusion and Limitations}
\label{sec:conclusion}

We introduce CycleSAM, a powerful few-shot SAM adaptation approach for surgical image segmentation.
Given a single labeled reference image, CycleSAM robustly predict point prompts for each object in a query image, then uses an off-the-shelf SAM to predict segmentation masks.
By leveraging domain-specific SSL-pretraining, CycleSAM effectively bridges domain gap, while a carefully designed training phase helps refine the SSL features and distinguish complex anatomical boundaries, a unique challenge in surgical images.
Through numerous experiments, we demonstrate that while existing few-shot SAM adaptations frequently fail on surgical images, CycleSAM achieves consistently strong performance. 
Crucially, we show that a specialist model trained on CycleSAM-generated pseudo-masks can reach up to 80\% of fully-supervised performance, demonstrating CycleSAM's potential to drastically reduce annotation costs.
A key limitation of CycleSAM is its complete reliance on SAM for mask decoding; though this can be improved by incorporating improved medical SAM adaptations in the future, it lacks filtering mechanisms to maximize mask quality.
Additionally, we only briefly explore CycleSAM's scaling capabilities, both with reference to increasing amounts of labeled data and as a pseudo-mask generator.
Analyzing performance when training CycleSAM on additional reference images as well as scaling the number of unlabeled images used for self-training is crucial to understanding how CycleSAM can be integrated into real-world modeling pipelines.